\documentclass[times, 10pt,twocolumn]{article} 
\usepackage{latex8}
\usepackage{times}
\usepackage{amsmath}
\usepackage[T1]{fontenc}
\usepackage[dvipdfmx]{graphicx}

\begin{document}

\title{An Efficient Method of Training Small Models for Regression Problems with Knowledge Distillation}

\author{Makoto Takamoto, Yusuke Morishita,  \\
Biometrics Research Laboratories, NEC \\ 
1753 Shimonumabe, Nakahara-ku, Kawasaki-shi, Kanagawa, Japan\\ makoto.takamoto@nec.com, \\
\and
Hitoshi Imaoka\\
NEC Corporation\\ 
1753 Shimonumabe, Nakahara-ku, Kawasaki-shi, Kanagawa, Japan\\
}

\maketitle
\thispagestyle{empty}

\begin{abstract}
  Compressing deep neural network (DNN) models becomes a very important and necessary technique
  for real-world applications, such as deploying those models on mobile devices. 
  Knowledge distillation is one of the most popular methods for model compression,
  and many studies have been made on developing this technique.
  However, those studies mainly focused on classification problems,
  and very few attempts have been made on regression problems,
  although there are many application of DNNs on regression problems. 
  In this paper, 
  we propose a new formalism of knowledge distillation for regression problems.
  First, 
  we propose a new loss function, teacher outlier rejection loss, 
  which rejects outliers in training samples using teacher model predictions. 
  Second,
  we consider a multi-task network with two outputs: one estimates training labels
  which is in general contaminated by noisy labels; 
  And the other estimates teacher model's output 
  which is expected to modify the noise labels following the memorization effects. 
  By considering the multi-task network, 
  training of the feature extraction of student models becomes more effective, 
  and it allows us to obtain a better student model than one trained from scratch. 
  We performed comprehensive evaluation with one simple toy model: sinusoidal function, 
  and two open datasets: MPIIGaze, and Multi-PIE.
  Our results show consistent improvement in accuracy 
  regardless of the annotation error level in the datasets. 
\end{abstract}


\Section{Introduction}

Recent development of deep neural network (DNN) research allows us to solve many kinds 
of problems with very high accuracy, such as classification, regression, and object detection.
To enhance DNNs' accuracy,
a straightforward method is just increasing depth and channel of the network,
and a considerable amount of studies have been conducted on finding a method to train
deeper networks effectively. 
Their large memory and numerical costs, however, prohibits us to apply them 
to real-world solutions. 
To alleviate this problem,
many techniques have been proposed, e.g., finding efficient network structures
~\cite{howard2017mobilenets,zhang2018shufflenet},
channel pruning \cite{han2015deep},
and quantization of network weights \cite{hubara2017quantized}. 
Knowledge distillation is one of the most popular methods for this purpose 
which tries to mimic the behavior of deeper and larger models (teacher) 
by a smaller or compressed model (student)
\cite{Bucilua:2006:MC:1150402.1150464,ba2014deep,hinton2015distilling}.
Although seminal work have followed to improve the technique, 
those work mainly focused on the classification problem. 
On the other hand, little attention has been given on regression problems 
which also have many applications, for example,
estimating age \cite{geng2007automatic}, gaze angle \cite{zhang2017mpiigaze}, and so on. 
Though some techniques of the above work can also be applied to regression problems, 
it has another inherent difficulty, that is, the uncertainty of giving annotation. 
In general, 
regression problems treat continuous variables as annotation, 
and it is unavoidable to accept a certain amount of annotation error 
which originates from human errors and limitations of measurement. 
Since information from teacher models can also be an origin of those errors, 
it is necessary to consider the treatment of these errors
when developing knowledge distillation for regression problems, 
which no existing work tackled as far as we know. 

To address the above problems,
we propose a method to train a fast and accurate student networks 
for regression problems using a newly developed knowledge distillation method. 
Our contributions are as follows:
\begin{itemize}
  \item 
    We propose a new formulation for knowledge distillation for regression problems 
    which solves regression problems using a multi-task network (Section \ref{sec:theory}). 
  \item We propose a new loss, so-called \textit{Teacher Outlier Rejection} (TOR) Loss 
    which allows the student models to avoid suffering from outliers 
    with the help of teacher models (Section \ref{sec:TOR}). 
  \item We present insights into the nature of regression problems with noisy data 
    which was obtained from comprehensive numerical experiments (Section \ref{sec:experiments}). 
\end{itemize}

\section{Related Work}

\textbf{Knowledge distillation} \quad Knowledge distillation is an approach of DNN model compression 
with retaining accuracy by trying transferring teacher model's knowledge to student models. 
After several pioneer work \cite{Bucilua:2006:MC:1150402.1150464,ba2014deep},
\cite{hinton2015distilling} proposed the temperature cross entropy loss
which regards the teacher prediction as a 'soft label',
expecting to transfer the teacher model knowledge about relation of each category. 
On the other hand,
\cite{adriana2015fitnets} claimed that imitating teacher's middle layer is also effective 
because it provides 'hint' to improve the training process and accuracy of student models. 

The above methods consider applications only to classification problems, 
and there are much smaller number of knowledge distillation methods for regression problems. 
A representative one is proposed by \cite{chen2017learning} 
which considered a knowledge distillation for object detection problems, 
and proposed a \textit{teacher bounded regression} (TBR) loss 
which tries to accelerate the student training until its error becomes less than teacher's error. 
In our method,
we used the teacher network output to find outliers which strongly affects the student training
and degrade its score,
and it is completely different from the above methods. 

\textbf{Regression on DNNs} \quad Regression problems appear in many situations, 
such as estimating age \cite{geng2007automatic}, gaze angle \cite{zhang2017mpiigaze},
body pose \cite{cao2018openpose}, and facial landmark points \cite{feng2018wing}. 
DNNs have been applied to solve those tasks, 
and many loss functions are proposed to improve its accuracy,
for example, wing-loss was proposed to accelerate the final convergence of facial landmark point estimation \cite{feng2018wing}. 
\cite{belagiannis2015robust} proposed the robust loss which tries to learn label data by rejecting outliers estimated
using the median absolute deviation (MAD) \cite{huber2011robust}. 

\textbf{Training on Noisy Data} \quad DNN training on noisy data is a hot topic.
\cite{li2017learning} proposed a framework to distill the knowledge from clean labels and knowledge graph. 
\cite{veit2017learning} proposed a multi-task network which learns both to clean noisy annotations and 
accurately classify images.
However, these methods demand a small dataset with clean labels. 
Another direction was proposed by \cite{han2018co} so called ``co-teaching'' technique. 
This technique uses the memorization effect \cite{Arpit:2017:CLM:3305381.3305406,zhang2016understanding}
which suggests that DNNs would first memorize training data of clean labels and then those of noisy labels. 
And this trains two DNNs simultaneously, and let them teach their estimated clean labels each other. 
The above methods consider only applications on classification problems,
and did not discuss applications on regression problems. 

\Section{\label{sec:theory}Theoretical Consideration}

\begin{figure}
 \center
   \includegraphics[width=7.5cm]{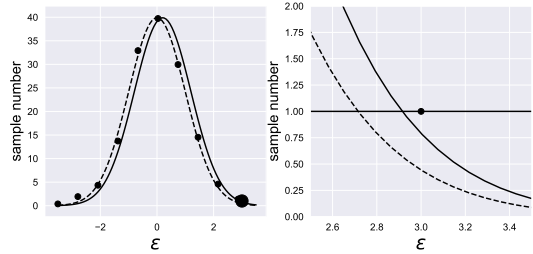}
   \caption{A schematic picture of the effect of outliers. 
     Left panel: whole picture of a distribution function of data samples in terms of error, 
     Right panel: around an outlier.
     The dashed line is the original line following: $100 \exp[-\epsilon^2/2]/\sqrt{2 \pi}$,
     points are the target points with small Gaussian noise,
     a bigger point is the outlier,
     and the solid line is the predicted line. 
     }
 \label{fig:outlier}
\end{figure}

In this section,
we present a theoretical consideration of our formalism. 
We consider a regression problem which tries to fit 1-dimensional data with noise $\epsilon$ 
which follows Gaussian distribution: $N(\epsilon|\mu,\sigma)$ with $\mu=0$. 
We assume that 
the training data is ${\bf x} = (x_1, \cdots, x_N)^T$,
and target data is ${\bf t} = (t_1, \cdots, t_N)^T$. 
From the above assumption on the error,
the probabilistic distribution of ${\bf t}$'s uncertainty can be written as: $p(t~|~x,w,\sigma) = N(t~|~y(x,w),\sigma)$ 
where $w$ is the parameter to be calculated by the fitting, 
and $y$ is the predicted function by our model.

To find a better $y$,
two strategies are considered in this paper.
The first one is to train a large teacher model, and learn its prediction as new labels.
As is reported by \cite{han2018co},
DNNs learn easy data first.
In the case of regression problems,
we expect that 
DNNs try to find the unknown true label and neglect noise $\epsilon$ because predicting all the noise is too hard for DNNs.
So, learning teacher prediction can be an easier task for small DNNs than learning original annotation.
As is reported in \cite{chen2017learning}, however,
this does not always work,
and learning original annotation is still fruitful for DNNs. 
This motivate us to develop another method 
which learns original annotation but robust to noise, 
and compensate the above defects 
by combining them as a multi-task network. 

\begin{figure}[t] 
 \center
   \includegraphics[width=4cm]{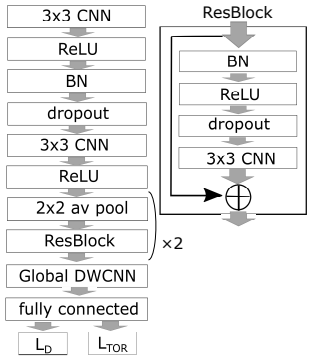}
   \caption{The network structure for image regression tests.
            In the figure,
            ``3x3 CNN'' means the convolution layer with filter size 3 and stride 1, 
            ``BN'' means the batch normalization layer, 
            ``2x2 av pool'' means the average pooling with size 2 and stride 2, 
            ``DWCNN'' means the global depthwise convolution,
              and ``$L_{\rm D}$'' and ``$L_{\rm TOR}$'' mean the outputs corresponding with
              ones learning teacher output and TOR prediction, respectively.
             }
 \label{fig:CNN}
\end{figure}

The second one is to consider the maximum likelihood estimation (MLE). 
It is well-known \cite{bishop2006pattern} that 
MLE of ${\bf w_{\rm ML}}$ is equivalent to the minimization
of mean-square error (MSE) loss function, 
which indicates that
moderate amount of noise in labels can be absorbed using MSE loss function 
as long as the noise follows the Gaussian distribution. 
On the other hand, 
it is also well-known that 
MSE is very vulnerable to outliers. 
Figure \ref{fig:outlier} shows an example of the effect of outlier. 
In the figure, 
the average of the center of estimated error curve $y(x,w)$ is deviated from the true error curve 
because of the existence of an outlier. 
This shows the significance of the impact on the estimation from outlier in ``tail-region'' of the probabilistic distribution, 
that is, even 1 sample gives strong effect if appearing in tail-region
where expectation of the appearance of the sample is less than a constant value $\alpha$ which is around unity:
\begin{equation}
  B~p(t_{\rm outlier}~|~x,w,\sigma) = B~N(t_{\rm outlier}~|~y(x,w),\sigma) < \alpha
  ,
  \label{eq:outlier}
\end{equation}
where $t_{\rm outlier}$ is the coordinate of the outlier, and $B$ is the sample number. 
Thus, 
it can be expected that
we can reduce the fitting error significantly 
if those outliers can be rejected
\footnote{
Practically speaking,
the noise of labels in databases can be expected to be the Gaussian because of the central limit theorem, 
so that the above assumption will be applicable in many cases. 
Note that, however,
Equation (\ref{eq:outlier}) is not limited to be the Gaussian but an arbitrary function $p(t~|~x,w)$,
and our loss function can easily be extended to them. 
}.
Note that such a sample in tail-region can always be appeared
because of their non-zero probability of existence even though the probability is very close to zero.
And it should be emphasized that
the optimization process is not i.i.d in the case of DNNs
because of the update of weight $w$ for each batch samples, 
and the effect from outliers cannot be erased even after many iterations. 
In the following sections, 
we provides an explanation 
how our loss function allows us to realize this statement. 

\begin{table}[t]
  \centering
  \caption{The threshold value ($\epsilon_{\rm outlier}$ and $\alpha$) dependence on the mean absolute error of a student network 
    obtained using TOR loss. 
    }
  \begin{tabular}{c|c|c} \hline
    $\epsilon_{\rm outlier}$ & student error   & expectation value ($\alpha$) \\ \hline    
    6 & 0.105 $\pm$ 0.014 & 4.5 \\
    7 & 0.099 $\pm$ 0.014 & 2.19 \\
    {\bf 8} & {\bf 0.092 $\pm$ 0.016} & {\bf 0.95} \\
    9 & 0.095 $\pm$ 0.018 & 0.37 \\ \hline
  \end{tabular}
  \label{table:table0}
\end{table}

\section{Method}

In this section,
we provide our new formalism of knowledge distillation for regression problems.
It solves regression problems using a multi-task network,
which is trained using several loss functions, 
in particular, those robust for noisy data in the case of regression problems. 
The effectiveness of each method will be discussed in Section \ref{sec:experiments}. 

\subsection{\label{sec:TOR}Outlier Rejection via Knowledge Distillation}

Different from classification problems,
it is unclear which information of teacher networks is fruitful for student networks
in the case of regression problems. 
In addition, \cite{chen2017learning} pointed out that
the information from teacher network may be even harmful 
if the teacher prediction is contradictory to the ground truth value.
Relating to this problem, 
recent studies \cite{zhang2016understanding,Arpit:2017:CLM:3305381.3305406} claimed a very interesting finding that 
DNNs learn from easy and clean label data first, 
and gradually remember remaining data.
This suggests that 
teacher networks can be used for finding noise data, 
and we propose the following loss function which we call \textit{Teacher Outlier Rejection (TOR) Loss}: 
\begin{align}
  L_{\rm TOR} &\equiv
  \begin{cases}
    ||R_{\rm s} - t||_2^2, & (\mathrm{if} \ |t - R_{\rm t}| < \epsilon_{\rm outlier}) 
    \\
    f(R_{\rm s} - R_{\rm t}), & (\mathrm{if} \ |t - R_{\rm t}| > \epsilon_{\rm outlier})
  \end{cases}  
\end{align}
where $t$ is the target labels, 
$R_{\rm s}, R_{\rm t}$ are the prediction of student and teacher models, respectively, 
$\epsilon_{\rm outlier}$ is a threshold judging if data is noise or not, 
and $f$ is an arbitrary function.
In this paper,
we set $f(x) = \sqrt{x}$ to reduce effects from noise data, 
although it is also possible to use other functions such as just setting 0. 
To estimate the threshold $\epsilon_{\rm outlier}$,
we use Equation (\ref{eq:outlier}).
By setting $\epsilon_{\rm outlier} \equiv |t_{\rm outlier} - y(x,w)|$ and
equating the left-hand side of Equation (\ref{eq:outlier}) to $\alpha$, 
it can be solved as: 
\begin{equation}
  \epsilon_{\rm outlier}= \sigma \sqrt{- 2 \ln \left[\sqrt{2 \pi} \sigma \alpha/B \right]}
  ,
  \label{eq:outlier_est}
\end{equation}
where $\alpha$ is a hyper-parameter around unity which defines the tail-region. 
Unfortunately,
the true $\sigma$ of general dataset is usually unknown, 
and we estimated it as $\sigma = 1.4826 \ {\rm MAD}$ 
where ${\rm MAD} \equiv {\rm median} |\xi - \mathrm{median} (\xi)|$ is the median absolute deviation, 
estimating a typical scale of the distribution function, 
and $\xi = t - R_{\rm t}$ \cite{huber2011robust}.

\subsection{\label{sec:multi}Multi-task Networks}

In our formalism,
we consider a multi-task network with 2 output: 
one estimates the label value without outliers, 
and the other estimates the teacher network prediction
which is expected to modify noise data. 
By considering multi-task networks, 
it can be expected that 
student models develop a good feature extraction part. 
In addition,
it can also be expected that
we can reduce statistical fluctuation by taking the average of those 2 outputs as a final output. 
To train our multi-task network, 
the following loss function was considered: 
\begin{equation}
  L = c_{\rm TOR} L_{\rm TOR} + c_{\rm D} L_{\rm D}
  ,
\end{equation}
where $c_{\rm TOR}, c_{\rm D}$ are numerical coefficients. 
$L_{\rm D}$ is the loss function to learn teacher prediction. 
In this paper,
we consider L1 loss for $L_{\rm D}$ but other loss functions can also be applied.

\begin{table*}[t]
  \centering  
  \caption{
    Results of the sinusoidal function regression test ($\times 10^2$).
    Note that the below value are multiplied by $10^2$ to reduce the table's size. 
    The baseline is the result of student model training with L1 loss function. 
    The listed scores were measured by the mean absolute error.
    The each test was performed 100 times, 
    and its average and the standard deviation were shown.
  }
  \begin{tabular}{l|c|c|c||c|c|c||c|c} \hline
    std & teacher & student (L1)        & student (MSE)        & {\bf ours full}     & {\bf only $L_D$} & {\bf only $L_{\rm TOR}$} & L1+TBR \cite{chen2017learning} & Robust Loss \cite{belagiannis2015robust} \\ \hline    
    0   & 2.2     & {\bf 6.9 $\pm$ 1.3} & 7.2 $\pm$ 1.4       & 7.0 $\pm$ 1.2       & 7.8 $\pm$ 1.2    & {\bf 6.9 $\pm$ 1.2} & 7.2 $\pm$ 1.5 & 8.1 $\pm$ 2.2 \\
    0.5 & 1.8     & 7.1 $\pm$ 1.2       & 7.3 $\pm$ 1.2       & {\bf 7.0 $\pm$ 1.2} & 7.3 $\pm$ 1.2    & 7.2 $\pm$ 1.1       & 7.3 $\pm$ 1.2 & 7.2 $\pm$ 1.4 \\
    1   & 2.7     & 7.4 $\pm$ 1.2       & 7.5 $\pm$ 1.2       & 7.5 $\pm$ 1.3       & 7.8 $\pm$ 1.2    & {\bf 7.3 $\pm$ 1.2} & 7.6 $\pm$ 1.1 & 7.7 $\pm$ 1.3 \\
    3   & 3.6     & 9.1 $\pm$ 1.3       & 8.4 $\pm$ 1.4       & {\bf 8.0 $\pm$ 1.3} & 9.4 $\pm$ 1.2    & 8.4 $\pm$ 1.4       & 9.1 $\pm$ 1.3 & 10.1 $\pm$ 1.3 \\
    5   & 5.0     & 9.7 $\pm$ 1.5       & {\bf 8.3 $\pm$ 1.5} & 8.4 $\pm$ 1.5       & 9.4 $\pm$ 1.2    & {\bf 8.3 $\pm$ 1.4} & 9.5 $\pm$ 1.2 & 17.8 $\pm$ 1.7 \\ \hline
  \end{tabular}
  \label{table:table1}
\end{table*}

\Section{\label{sec:experiments}Experiments}
In this section,
we provide the results of our experiments for knowledge distillation  
for regression problems. 
We added a small Gaussian noise to the label data
in order to simulate regression problems on noisy data 
and to confirm the effectiveness of our new formalism. 
In the following, 
explanations of the used datasets and models are provided.
In Section \ref{sec:sin},
results of fitting the sinusoidal function is explained. 
In Section \ref{sec:CNN},
the applications of our method to more difficult problems
predicting gaze direction and head-pose are performed.
In those tests,
we performed many trainings, 100 trials for sinusoidal function and 10 trials for image datasets, 
to reduce statistical fluctuations resulting from using small models.
Concerning the tests on image datasets,
we prepared 10 initial weights of the student model for each standard deviation 
and used them for all the tests with the same standard deviation 
to reduce the effect of the initial weight selection. 

\textbf{Datasets} \quad We evaluate our method using several datasets. 
First, 
we consider the sinusoidal function with noise as a simple toy problem. 
The data size is $10^5$,
and the noise is the Gaussian noise with various standard deviation (std) value. 
Although this is a very simple problem,
this provides us with an important insight of our method on noisy data. 
Next, we used the MPIIGaze dataset \cite{zhang2017mpiigaze} 
which is one of the most popular dataset for gaze estimation,
including more than $4 \times 10^5$ gazing data with annotation.
We used $2 \times 10^4$ data randomly selected from this database for each training. 
The final one is the CMU Multi-PIE Face Database \cite{Gross:2010:MUL:1746745.1747071} 
which contains about $1.5 \times 10^4$ images of faces taken from 15 view points 
for the head-pose estimation,
so that this is a kind of classification problem of head-pose with soft-labels. 
We used $10^4$ data randomly selected from the session1 of this database for each training. 

We transformed the image data into gray scale, 
and normalized them by subtracting mean value and dividing by the standard deviation value of each image.
Concerning Multi-PIE, 
the resolution of the image data were down-sampled from $640 \times 480$ to $80 \times 60$. 
We did not crop the images around face for simplicity. 

\textbf{Models} \quad For the sinusoidal function regression test,
a multi-layer perceptron with 1 hidden layer was used for both of the teacher and student models. 
The input and hidden layers are followed by ReLU \cite{dahl2013improving}, 
Batch Normalization \cite{pmlr-v37-ioffe15}, and dropout layers \cite{srivastava2014dropout}. 
The teacher has 150 channels, and the student has 40 channels. 
Concerning the networks and training parameters, 
we set the batch size to 1000, 
dropout to 0.5,
total epochs to 100, 
and the learning rate to $10^{-3}$ 
which was dropped by 0.1 at 70 epochs for the student
and at 40 and 80 epochs for the teacher. 
The optimization was performed using Adam \cite{kingma2014adam}.

For the image regression tests,
we used a Wider Resnet-like network \cite{BMVC2016_87} 
whose structure is presented in Figure \ref{fig:CNN}.
In all the layer other than the 1st CNN layer with 40 channels, 
the teacher has 80 channels, 
and the student has either 20 or 40 channels and half-channel size 1st CNN layer. 
The detailed explanation of the channel number of student is listed in Table 3. 
Concerning the networks and training parameters, 
we set the batch size to 1024, 
dropout to 0.25, 
total epochs to 100, 
and the learning rate to $7.5 \times 10^{-4}$ 
which was dropped by 0.1 at 40 and 80 epochs. 
The optimization was performed using Adam. 
In this work, 
we did not use data augmentation to avoid adding another noise source. 

\begin{table*}[t]
  \begin{center}
  \caption{
    Results of the image regression tests. 
    The listed scores were measured by the mean absolute error. 
    The each test was performed 10 times, 
    and its average is shown.
    (The full result including the Robust Loss \cite{belagiannis2015robust} is provided in the published version)
  }
  \begin{tabular}{l|c|c|c|c|c||c|c|c||c|c} \hline
    DataSet & std & channel & teacher & student (L1) & student (MSE) & {\bf ours full} & {\bf only $L_{\rm D}$} & {\bf only $L_{\rm TOR}$} & L1+TBR \cite{chen2017learning} \\ \hline \hline
    MPIIGaze  & 0  & 20 & 1.171 & 1.648 & 1.664 & 1.621       & {\bf 1.620} & 1.654 & 1.640 \\ 
    MPIIGaze  & 2.5& 20 & 1.237 & 1.665 & 1.680 & {\bf 1.630} & 1.642       & 1.669 & 1.668 \\
    MPIIGaze  & 5  & 20 & 1.383 & 1.712 & 1.714 & 1.669       & {\bf 1.664} & 1.718 & 1.710 \\     
    MPIIGaze  & 10 & 20 & 1.764 & 1.838 & 1.802 & {\bf 1.729} & 1.748       & 1.804 & 1.832 \\
    MPIIGaze  & 5  & 40 & 1.381 & 1.503 & 1.482 & {\bf 1.453} & 1.484       & 1.485 & 1.503 \\ \hline       
    Multi-PIE & 5  & 20 & 1.296 & 1.899 & 1.978 & {\bf 1.869} & 1.919       & 1.960 & 1.945 \\
    Multi-PIE & 10 & 20 & 2.009 & 2.186 & 2.201 & {\bf 2.095} & 2.179       & 2.188 & 2.191 \\ \hline      
  \end{tabular}
  \end{center}  
  \label{table:table2}
\end{table*}

\subsection{\label{sec:sin}Sinusoidal Function Regression Tests}
First,
we discuss the validity of the definition of outliers 
discussed in Section \ref{sec:TOR}. 
Table \ref{table:table0} presents the threshold value ($\epsilon_{\rm outlier}$ and $\alpha$ in Equation (\ref{eq:outlier_est})) dependence
of the mean absolute error of student network 
obtained using TOR loss.
In this test, the added noise std is 3, and the small size dataset (10000 samples) was used with batch size 250. 
Base line error (only L1 loss) is: $0.112 \pm 0.016$.
The each test was performed 100 times, 
and its average and the standard deviation were shown. 
It shows that 
the error gradually reduces as decreasing the expectation value, which corresponds with $\alpha$.
Note that this behavior is natural
because too small $\alpha$ results in a very close result to MSE loss which is vulnerable to outliers, 
and too large $\alpha$ reduces the number of the less noisy fruitful data. 
Importantly,
the error reaches an lower limit when the expectation value becomes around unity, 
which shows the validity of our assumption that
there is an threshold of defining outliers whose $\alpha$ is around unity as we set in Section \ref{sec:TOR}
\footnote{
  Our numerical experiments using other hyper-parameters, such as batch number and the standard deviation,
  showed a similar behavior of the student error dependence on $\alpha$, 
  although sometimes showing more fluctuated results because of the randomly chosen initial weight.
  Those experiment results indicates that
  the best $\alpha$ can be found in between from 0.1 to 10,
  and in many case around unity. 
}.

Table \ref{table:table1} is a summary of the test results.
The coefficients of losses are set as: $(c_{\rm TOR}, c_{\rm D}) = (1, 1)$
in the case of std=0,~0.5,~5, and $(c_{\rm TOR}, c_{\rm D}) = (10, 1)$ for the others. 
The $\alpha$ is set as unity in the case of std=0,~3, and 2 for the others. 
It shows that 
our full-result (multi-task case) provides the lowest error for almost all the cases.
When using only $L_{\rm D}$ loss, 
it shows a little larger error for all the cases comparing with the other method.
We consider that this is because the teacher prediction always deviates from the true solution,
and this difference results in the asymmetric distribution of the noisy labels.
However, we emphasize that this setup (perfect label on regression) is unrealistic
because of the intrinsic difficulty of regression problem to give perfect annotation. 
Concerning the case only using $L_{\rm TOR}$, 
we found that
this loss becomes comparable to L1 loss and better than MSE loss in the small std region.
On the other hand, 
it becomes better than L1 loss and comparable to MSE loss in the large std region. 
We consider that
this is because in the large std region MSE loss is more consistent with the maximum likelihood estimation
of Gaussian noise than L1 loss;
Unexpectedly, the errors of the case only using TOR loss are comparable to those using MSE loss
in the large std region.
This may be because the too strong noise makes the training too difficult, 
and the effect of rejecting outliers becomes negligible
\footnote{
Note that
our full-result sometimes becomes worse than the cases only using either $L_{\rm D}$ or $L_{\rm TOR}$. 
This is related to the selection of the coefficients of losses $c_{\rm TOR}, c_{\rm D}$. 
}. 
In all ranges, 
both L1+TBR loss \cite{chen2017learning} and robust loss \cite{chen2017learning} work well but 
worse than the baseline and our method. 
We found that
L1+TBR loss basically shows a very similar ones to the case only considering L1 loss, 
and Robust loss shows large standard deviation. 

\subsection{\label{sec:CNN}Tests on Image Datasets}

In this section,
test results using image data were provided.
The coefficient of losses are set as: $(c_{\rm TOR}, c_{\rm D}) = (1, 25)$ for the cases of $\mathrm{std}=0, 2.5$ of MPIIGaze,
and $(c_{\rm TOR}, c_{\rm D}) = (1, 10)$ for all the other cases. 
Table 3 is a summary of the test results. 
For MPIIGaze tests,
the results were averaged for pan and tilt gaze angle.
For Multi-PIE tests,
only the results for pan angle were shown because it has only 2 tilt angles
\footnote{
  Here we did not perform Multi-PIE test with std=0. 
  This is because it is more close to a classification problem as pointed out,
  and cannot be an appropriate test for revealing the accuracy
  of our new technique on regression problems with intrinsic annotation errors. 
}. 

The tendency of the results is similar to that of the sinusoidal function regression given in Section \ref{sec:sin}. 
Our full-result shows the best results in almost all the cases,
showing the effectiveness of considering multi-task networks. 
Interestingly, 
it shows a slightly better result than the baseline (L1, MSE) even when no noise was added, 
indicating the fundamental difficulty to erase all the annotation error even for popular database
which are carefully given their annotation.
TOR loss shows better results than MSE loss in the case of the small std cases,
and similar ones in the case of the large std cases.
Different from the sinusoidal function test, 
the case only considering $L_{\rm D}$ shows either the second or the best results in most cases, 
which indicate that
the teacher network can estimate the true label of the dataset as is discussed above. 
This is also implied from the results of only considering $T_{\rm TOR}$ in the no noise case (std$=0$) 
which is better than the MSE loss case by reducing the contribution from outliers. 
In the case of giving large noise (std=10) for MPIIGaze dataset,
our method provided a better score than the teacher models.
We consider that
this is because the smaller capacity of student model reduced the effect of over-fitting of the teacher models. 

\Section{Discussion and Conclusion}

In this paper,
we proposed a new formalism of knowledge distillation for regression problems,
which used teacher models as an new annotator an outlier detector. 
Our experiments showed that
the model with sufficiently large amount of parameters can learn training data
without being annoyed by noise,
and can be used for outlier detector. 
The results also showed that 
our knowledge distillation method can be effective
even when large noise is added as is expected.
Concerning learning teacher prediction,
our experiment showed that
it is very effective when the label is contaminated. 
Concerning the TOR loss,
we found that
its prediction is slightly worse than L1 and better than MSE when noise is small; 
And it is better than L1 and equivalent to MSE when noise is large
\footnote{
  In practice, it will be a good option to add L1 prediction to the multi-task network output,
  which we avoid in this paper for simplicity. 
}. 
Note that 
in actual use cases 
the collected data with hand-crafted label usually include a lot of noise, or incorrect labels,  
which is fundamental for regression problems, and nearly impossible to reduce to 0. 
Our method allows to avoid being affected from such unavoidable error in annotation,
and to train a better DNN model.

%
%

\bibliographystyle{latex8}

\end{document}